\documentclass[11pt]{article}
\usepackage[margin=1in]{geometry}
\usepackage{graphicx}
\usepackage{amsmath,amssymb,amsfonts}
\usepackage{cite}
\usepackage{hyperref}
\usepackage{algorithmic}
\usepackage{textcomp}

\title{
X2BR: High-Fidelity 3D Bone Reconstruction\\ 
from a Planar X-Ray Image\\ 
with Hybrid Neural Implicit Methods
}

\author{
Gokce Guven$^{1}$, H. Fatih Ugurdag$^{1}$, Hasan F. Ates$^{1}$ \\
$^{1}$Faculty of Engineering, Ozyegin University, Istanbul, Turkey\\
\texttt{\{gokce.guven, fatih.ugurdag, hasan.ates\}@ozu.edu.tr}
}

\date{}

\begin{document}

\maketitle

\begin{abstract}
Accurate 3D bone reconstruction from a single planar X-ray remains a challenge due to anatomical complexity and limited input data. We propose X2BR, a hybrid neural implicit framework that combines continuous volumetric reconstruction with template-guided non-rigid registration. The core network, X2B, employs a ConvNeXt-based encoder to extract spatial features from X-rays and predict high-fidelity 3D bone occupancy fields without relying on statistical shape models. To further refine anatomical accuracy, X2BR integrates a patient-specific template mesh, constructed using YOLOv9-based detection and the SKEL biomechanical skeleton model. The coarse reconstruction is aligned to the template using geodesic-based coherent point drift, enabling anatomically consistent 3D bone volumes. Experimental results on a clinical dataset show that X2B achieves the highest numerical accuracy, with an IoU of 0.952 and Chamfer-L1 distance of 0.005, outperforming recent baselines including X2V and D2IM-Net. Building on this, X2BR incorporates anatomical priors via YOLOv9-based bone detection and biomechanical template alignment, leading to reconstructions that, while slightly lower in IoU (0.875), offer superior anatomical realism, especially in rib curvature and vertebral alignment. This numerical accuracy vs. visual consistency trade-off between X2B and X2BR highlights the value of hybrid frameworks for clinically relevant 3D reconstructions.

\end{abstract}


\section{Introduction}
Neural implicit representations have revolutionized 3D shape modeling by enabling high-fidelity reconstructions without explicit parameterization. These methods encode objects as continuous functions, allowing for precise 3D reconstruction, novel view synthesis, and shape interpolation. Their ability to model fine geometric details makes them essential for applications in virtual reality, robotics, and medical imaging.

Single-image 3D reconstruction, powered by deep learning, has significantly advanced spatial understanding from 2D inputs, benefiting fields such as augmented reality, autonomous navigation, and medical diagnostics. In medical imaging, neural implicit representations provide a powerful tool for high-resolution volumetric reconstruction, enabling detailed anatomical modeling for applications such as surgical planning, biomechanical analysis, and patient-specific treatment design. However, their continuous and unconstrained volumetric outputs may not inherently preserve anatomical structure or biomechanical validity. Therefore, integrating neural implicit methods with domain-specific priors—such as biomechanical templates or anatomical landmarks—is essential to ensure anatomically consistent and clinically reliable reconstructions.

Recent advances demonstrate the efficacy of neural implicit methods across various medical domains. X2Teeth \cite{b1} reconstructs individual teeth from panoramic radiographs, while Oral-3Dv2 \cite{b2} employs implicit functions to map 2D coordinates to 3D dental structures. ToothInpainter \cite{b3} fuses partial 3D models and X-rays for comprehensive dental reconstructions, including roots. In broader medical imaging, MedNeRF \cite{b4} generates high-resolution CT-like projections from sparse X-rays, ImplicitVol \cite{b5} reconstructs 3D ultrasound volumes without voxel grids, and SAX-NeRF \cite{b6} applies line-based transformers for improved sparse-view X-ray reconstructions. SNAF \cite{b7} further extends these capabilities by refining CBCT reconstructions with neural attenuation fields. These studies highlight the transformative impact of implicit representations in medical imaging.

This study introduces X2B and X2BR, two complementary neural implicit frameworks for 3D skeletal reconstruction from a single planar X-ray. X2B employs a ConvNeXt-based encoder to extract hierarchical spatial features and predict continuous occupancy fields, enabling 3D reconstruction of complex skeletal structures such as ribs and vertebrae. It effectively handles challenges such as overlapping Hounsfield Unit (HU) values and incomplete anatomical input without relying on voxel grids or predefined statistical templates. Building upon this, X2BR integrates a template-guided non-rigid registration step using a biomechanical skeleton model and geodesic-based coherent point drift (GBCPD++). This hybrid approach refines the initial reconstruction, ensuring anatomical consistency and improving alignment to patient-specific skeletal variations. Together, X2B and X2BR offer powerful solutions for accurate and personalized 3D bone modeling from sparse imaging data.

Experiments on clinical datasets demonstrate that X2B achieves state-of-the-art accuracy, significantly outperforming existing methods in volumetric IoU, Chamfer-L1 distance, and F-score. X2BR, while yielding slightly lower numerical accuracy, provides significantly improved anatomical consistency by incorporating a biomechanical template into the reconstruction process. This template serves as a prior anatomical structure, guiding the non-rigid registration of the occupancy-based output. By aligning the reconstructed volume to a deformable skeletal model, X2BR enforces biomechanical plausibility and better accommodates patient-specific anatomical variations. The framework’s ability to generate high-resolution 3D skeletal reconstructions from sparse imaging data—while leveraging a deformable anatomical template—makes it particularly well-suited for applications such as surgical planning, orthopedic assessment, and patient-specific biomechanical simulations.

In summary, the contributions of this study are as follows:

\begin{itemize}
    \item Proposes a hybrid neural implicit framework for 3D skeletal reconstruction from a single planar X-ray, combining a template-free occupancy-based model (X2B) with a template-guided refinement module (X2BR) for enhanced anatomical consistency.
    \item Incorporates ConvNeXt-based encoder to enhance spatial feature extraction for precise skeletal reconstruction.
    \item Handles anatomical challenges, including overlapping HU values and missing anatomical regions.
    \item Enables high-resolution, continuous and anatomically consistent modeling of 3D bone structures.
    \item Introduces the largest real-patient dataset of 3D bone meshes and corresponding DRRs (digitally reconstructed radiograph).
    \item Achieves state-of-the-art performance, surpassing existing methods in IoU, Chamfer-L1, and F-score.
\end{itemize}

\section{Related Work}
\subsection{Single-View Reconstruction with Implicit Surface Representations}

Neural implicit methods commonly use MLPs to represent occupancy probabilities or signed distance functions (SDFs) for 3D reconstruction from single images \cite{b8, b9}. While previous neural implicit approaches effectively reconstruct general 3D shapes, they fail to address the anatomical complexity of skeletal structures from single planar X-rays; our proposed X2B and X2BR frameworks specifically bridge this gap, improving reconstruction accuracy for complex skeletal anatomies.

Building on these foundations, recent CNN-based methods such as DISN \cite{b9}, MDISN \cite{b10}, and Ray-ONet \cite{b11} improve reconstruction fidelity but continue to face limitations in accurately modeling complex geometries. More advanced frameworks like D2IM-Net \cite{b12}, ED2IF2-Net \cite{b13}, G2IFu \cite{b14}, and LIST \cite{b15} further enhance topological accuracy and surface detail reconstruction. However, these approaches are primarily designed for general-purpose 3D reconstruction and remain insufficient when applied to anatomically intricate medical structures from sparse, single-view X-ray data.

\begin{figure*}[h!]    
        \centering
        \includegraphics[scale=0.5]{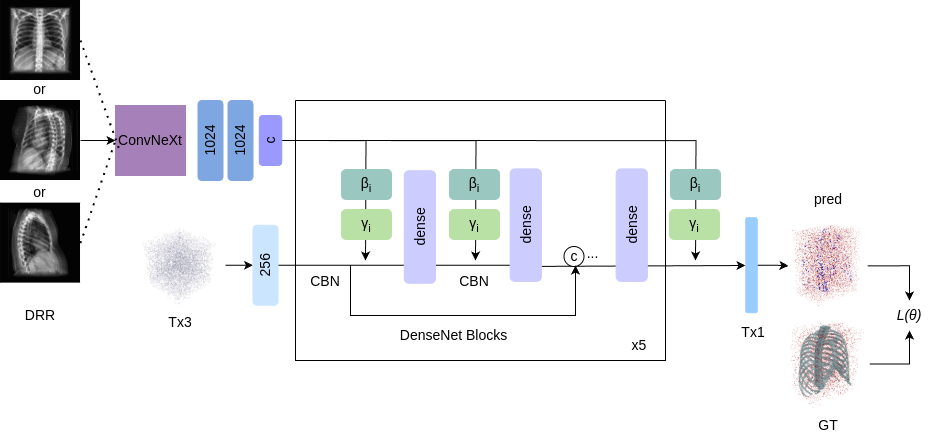}
        \caption{X2B network training pipeline. The figure illustrates the training process of the X2B network, which takes a DRR as input and uses a ConvNeXt backbone for feature extraction. The extracted features are passed through dense blocks with Conditional Batch Normalization (CBN) layers, parameterized by $\beta_i$ and $\gamma_i$, to refine the latent representations.}
        \label{guven1}
\end{figure*}
\subsection{X-ray to 3D Bone Reconstruction with Deformation Learning}

Statistical Shape Models (SSMs) utilize atlases derived from healthy samples to model mean anatomical shapes and variations \cite{b16,b17}. Aubert et al. \cite{b18} combined SSMs with CNN-based landmark detection for automated 3D spinal reconstruction. Jiang et al. \cite{b19} proposed a 2D/3D registration method for spinal geometry reconstruction from frontal X-rays, while X23D \cite{b20} integrated multi-view stereo and X-ray calibration for intraoperative vertebrae modeling.

More recent deep learning approaches also leverage deformation modeling. BX2S-Net \cite{b21} employs bi-planar X-rays with encoder-decoder architectures and attention mechanisms for improved semantic alignment. Similarly, Yang et al. \cite{b22} adapted X2CT-GAN \cite{b23} to reconstruct spinal structures from bi-planar radiographs. For single-view reconstruction, the approach in \cite{b24} utilizes deep learning with deformation parameters to reconstruct accurate 3D femoral models, while FracReconNet \cite{b25} improves fracture reconstruction accuracy by augmenting training data.

Existing methods often depend on bi-planar inputs, explicit deformation models, or statistical shape templates, limiting their generalizability and patient specificity. In contrast, X2B reconstructs high-fidelity 3D skeletal structures directly from single planar X-rays without requiring such priors. X2BR further enhances anatomical consistency by integrating a patient-specific template and geodesic-based non-rigid alignment. Together, they combine the strengths of data-driven reconstruction and anatomy-aware refinement for robust single-view bone modeling.

\subsection{Implicit Neural Representations in Medical Imaging}

Recent advances in implicit neural representations have significantly improved medical imaging reconstruction. X2Teeth \cite{b1} employs three subnets—ExtNet, SegNet, and ReconNet—to extract features, enhance segmentation, and reconstruct individual teeth. Oral-3Dv2 \cite{b2}maps 2D coordinates to 3D voxel densities, leveraging dynamic sampling for improved detail. ToothInpaintor \cite{b3} reconstructs full dental models, including roots, using implicit representations. MedNeRF \cite{b4} employs neural radiance fields to generate CT projections from X-rays, while ImplicitVol \cite{b5} refines 3D volumes from 2D ultrasound without voxel grids. SAX-NeRF \cite{b6} and SNAF \cite{b7} enable 3D reconstruction from sparse X-ray views using specialized sampling and augmentation techniques.  Our previous work X2V \cite{b26} utilizes a ViT-based implicit model to reconstruct 3D lung volumes from a single X-ray, without relying on mesh templates, and achieves state-of-the-art accuracy using occupancy networks.

Despite these advances, most existing methods are constrained to soft tissues, isolated anatomical regions, or require regular, high-contrast geometries. For instance, X2V is limited to organ volumes like lungs and relies on air-filled structures for accurate occupancy estimation. In contrast, our X2BR framework extends implicit neural representations to complex skeletal anatomy—including articulated structures such as vertebrae—using biomechanical template alignment and non-rigid registration. This enables anatomically consistent, high-fidelity 3D reconstructions from a single planar X-ray, even in the absence of dense multi-view inputs or statistical shape priors.

\section{Proposed Method}

Reconstructing accurate 3D bone structures from single planar X-rays is challenging due to anatomical complexity and overlapping tissue intensities. This study introduces two neural implicit models, X2B and X2BR, for precise and patient-specific 3D reconstructions.

\subsection{X2B Network Architecture}

The X2B network processes a \(224 \times 224\) DRR image and \(T\) random 3D points to predict occupancy probabilities, indicating whether a point lies within the bone surface (see Figure \ref{guven1}). A ConvNeXt-based encoder extracts hierarchical spatial features, transforming the input into a 1024-dimensional latent representation. ConvNeXt optimizes computational efficiency using depthwise convolutions, GELU activations, and Layer Normalization (LN). The occupancy network combines ConvNeXt features with 3D points, processed through DenseNet blocks with Conditional Batch Normalization (CBN) \cite{b27}, which dynamically adjusts normalization based on contextual features. This architecture ensures robust feature extraction and smooth reconstruction, leveraging the occupancy function \cite{b8} to implicitly define the bone surface as a decision boundary. As explained below, DRRs captured from multiple angles are used during training of X2B to enhance the network’s ability to reconstruct 3D structures with greater precision.

\begin{figure*}[!h]    
        \centering
        \includegraphics[scale=0.4]{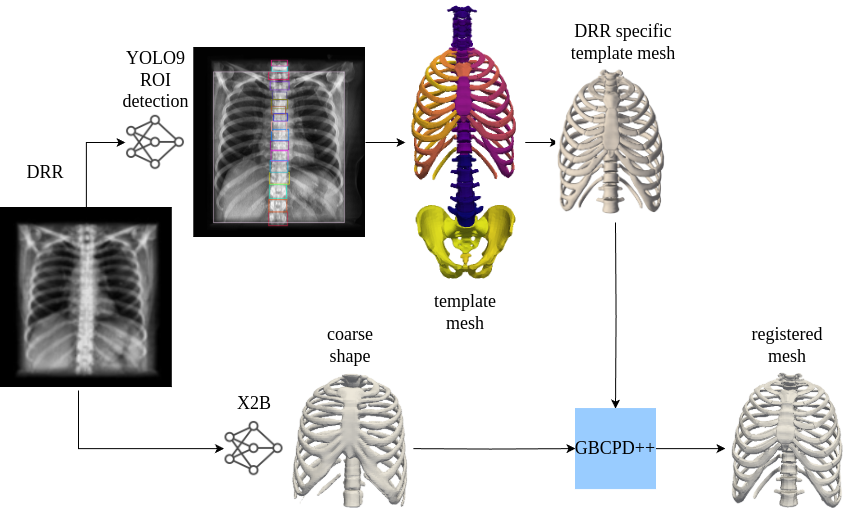}
        \caption{X2BR model architecture. Anterior-posterior DRR is used as input for both YOLOv9 and X2B for inference. The regions are detected via YOLOv9 network and DRR specific template mesh is extracted from the template mesh model using the detected regions. DRR specific template model is registered to the course shape, which is the output of the X2BR model.}
        \label{guven2}
\end{figure*}

\subsection{X2BR Network Architecture}

Building on X2B, X2BR integrates non-rigid registration to refine coarse reconstructions (see Figure \ref{guven2}). A fine-tuned YOLOv9 model identifies vertebrae (cervical, thoracic, lumbar) and ribs from DRR images. These detected regions are combined with the SKEL Biomechanical Skeleton Model (BSM) \cite{b28}, developed using OpenSim \cite{b29}, to construct a patient-specific template mesh. The Geodesic Bayesian Coherent Point Drift (GBCPD++) algorithm \cite{b27} aligns the coarse 3D shape from X2B with the patient-specific template, addressing complex deformations. 

The combination of ConvNeXt encoding, CBN normalization, YOLOv9 object detection, and GBCPD++ alignment enables X2B and X2BR to address the limitations of traditional methods. These models deliver precise 3D reconstructions, making them suitable for applications in surgical planning, biomechanical analysis, and personalized treatment.
\begin{figure*}[ht]    
        \centering
        \includegraphics[scale=0.4]{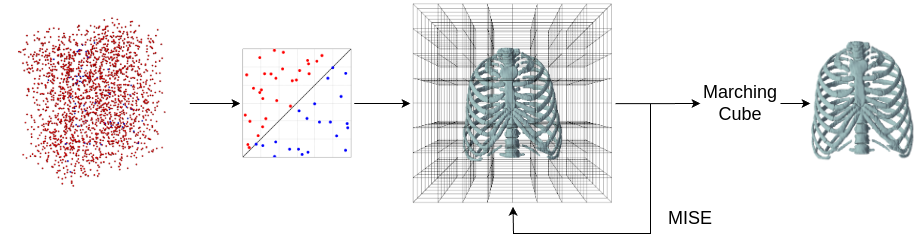}
        \caption{For 3D mesh inference with \textit{X2B} model, a modified Multiresolution IsoSurface Extraction (MISE) algorithm for high-resolution mesh extraction is integrated, starting with a base resolution and evaluating against the occupancy network. The occupancy threshold is set at $\tau = 0.2$ for balance in accuracy and completeness. The process involves subdividing voxels until the desired resolution is reached, using Marching Cubes for mesh generation, and refining the mesh with Fast-Quadric-Mesh-Simplification and gradient optimization. Our method achieves efficient and accurate mesh inference, optimized for an initial resolution of $32^3$, and is capable of extracting mesh normals effectively.}
        \label{guven3}
\end{figure*}

The SKEL Body Shape Model (BSM) provides watertight meshes of the thorax, pelvis, and spine, serving as the foundation for generating a DRR-specific template model. The X2BR model utilizes these meshes to construct subject-specific skeletal templates optimized for non-rigid registration. By leveraging the SKEL BSM, the X2BR model ensures high anatomical fidelity while efficiently adapting the template to align with input DRR data. This enables precise deformation and anatomically accurate reconstructions tailored to individual subjects.

\subsection{Training of X2B}

The X2B network is trained to reconstruct 3D skeletal structures from DRRs. As depicted in Figure \ref{guven1}, during training of the model a randomly selected input DRR image that is captured at different angles is passed through a ConvNeXt backbone, which extracts multi-dimensional feature representations. These features are then processed through a series of DenseNet blocks that include Conditional Batch Normalization (CBN) layers, parameterized by $\beta_i$ and $\gamma_i$, to refine the latent representations. The network iteratively processes these features through multiple dense layers, producing a final prediction that reconstructs the 3D skeletal structure. The loss function $L(\theta)$ is computed by comparing the predicted 3D output to the ground truth, enabling the network to optimize its parameters for accurate reconstruction.

The training objective is to estimate the occupancy at every point \( p \in \mathbb{R}^3 \) to derive the coarse shape from the X2B network's output. The occupancy function 
\[ o : \mathbb{R}^3 \rightarrow \{0, 1\} \]
defines whether a 3D point is occupied. This function can be approximated by a neural network, which assigns each point \( p \) an occupancy probability between 0 and 1, using a binary classification network's decision boundary to implicitly represent the object's surface.

The network, represented as \( f: \mathbb{R}^3 \times X \rightarrow \mathbb{R} \), maps the pair \( (p, x) \) to a real number indicating the probability of occupancy, where \( x \in X \) is the observation conditioning the 3D reconstruction (3Dr) task.

Training the binary-classification network involves computing the mini-batch loss \( L_B(\theta) \) at randomly sampled points within the object's 3D bounding volume. This is expressed as:
\begin{equation}
L_B(\theta) = \frac{1}{|B|} \sum_{i=1}^{|B|} L(f_\theta(p_{i}, x_i), o_{i})
\label{eq:2}
\end{equation}
where \( x_i \) is the \(i^{th}\) observation in batch \( B \), \( o_{i} \equiv o(p_{i}) \) represents the true occupancy at point \( p_{i} \), and \( L(\cdot , \cdot) \) is the binary cross-entropy classification loss. The method's effectiveness depends on the sampling strategy for locations \( p_{i} \) during training, with optimal results achieved through uniform sampling within the object's bounding box \cite{b8}.

For all experiments, the Adam optimizer \cite{b30} was employed with a learning rate (\(\eta\)) of \(10^{-4}\). The default hyperparameters provided by PyTorch were used: \(\beta_1 = 0.9\), \(\beta_2 = 0.999\), and \(\epsilon = 10^{-8}\).

\subsection{Inference}

Figure \ref{guven3} demonstrates the inference of 3D meshes from occupancy predictions at the output of the X2B network. In the X2B implementation, a modified Multiresolution IsoSurface Extraction (MISE) algorithm \cite{b8} is employed to enable efficient high-resolution mesh extraction.

The process starts with the discretization of the volumetric space and the evaluation of occupancy values at this initial resolution. Voxelization involves thresholding to classify each voxel as inside or outside the object based on the model's predictions. This critical step converts probabilistic outputs into a discrete 3D representation, facilitating accurate reconstruction from a single image viewpoint. Grid points exceeding an occupancy threshold, set to \(\tau = 0.2\) for the X2B method, determine the surface thickness, balancing accuracy and completeness as per ONet recommendations \cite{b8}.

Active voxels are identified and subdivided iteratively until the desired resolution is reached. The Marching Cubes algorithm then generates the initial mesh, which is refined using the Fast-Quadric-Mesh-Simplification algorithm. This algorithm applies iterative edge contraction and quadric error metrics, followed by gradient-based optimization, ensuring mesh quality enhancement and simplification.

\subsection{Registration}

The Geodesic-Based Bayesian Coherent Point Drift algorithm \cite{b27} introduces a non-rigid registration method addressing a significant drawback of the traditional Coherent Point Drift (CPD) algorithm. CPD, while prevalent for shape matching and deformation, often unnaturally deforms shapes with neighboring parts, such as human legs. This issue arises from the proximity-based deformation constraint known as motion coherence. The GBCPD++ method redefines motion coherence using geodesic distances, the shortest paths on a shape's surface, to mitigate inappropriate deformations. Numerical experiments demonstrate the efficacy of the geodesic-based approach in avoiding CPD's shortcomings and its scalability to handle shapes with millions of points. Key contributions include utilizing geodesic and Gaussian kernels for improved registration, providing a theoretical basis for converting indefinite geodesic kernels into positive-semidefinite ones, and employing the Nyström method and parallel computations to accelerate the process. The GBCPD++ method, integrated into the X2BR model, refines the reconstruction of 3D skeletal structures by enabling anatomically consistent registration of detailed vertebral features. By leveraging geodesic distances and scalable computations, it addresses complex deformations and enhances the accuracy of skeletal alignment.

\section{Dataset}

The dataset curated for this study is essential for advancing accurate 3D bone reconstruction from X-ray images. Due to the lack of publicly available paired datasets of X-rays and 3D bone meshes, a custom, high-quality dataset was created. Using advanced segmentation techniques, state-of-the-art DRR technology, and preprocessing pipelines, the dataset includes CT scans, DRR images, and 3D bone meshes systematically processed to ensure precision and suitability for training and validation. This section details the data acquisition, segmentation, DRR generation, and occupancy value computation, enabling efficient and accurate mesh reconstruction.

\subsection{Segmentation with TotalSegmentator}

The X2B network isolates organs from X-ray images, focusing on ribs, costal cartilages, and vertebrae due to their high HU contrast and the abundance of thorax data. Early experiments demonstrated the superior accuracy of TotalSegmentator \cite{b31} for bone segmentation, prompting its use in this study. Scapula bones were excluded due to frequent segmentation errors. Dataset generation steps are summarized as follows:

\begin{itemize}
  \item Data collection from The Cancer Imaging Archive (TCIA).
  \item Bone segmentation from CT images into watertight manifold meshes using TotalSegmentator \cite{b31}.
  \item Calculation of occupancy values for bone mesh points.
  \item Generation of corresponding DRR images from CT scans in multiple views.
  \item Enhancement of DRR images using CLAHE contrast adjustment.
\end{itemize}

CT scans from 964 subjects in the NLST dataset \cite{b32} were curated, excluding scans with incomplete thorax coverage or poor quality. DRR technology \cite{b33} was utilized to synthesize paired X-rays from real CT volumes. Scans were aligned to consistent anatomical planes, and low-quality segmentations were excluded. 

A total of 3,640 CT scans from 964 subjects were processed. Segmentation achieved a Dice Similarity Coefficient (DSC) of 0.943 ± 0.04 \cite{b31}. A custom Python script \cite{b34, b35, b36, b37} automated resampling to 1.5mm isotropic spacing, segmentation, mesh extraction, and merging of segmented regions into a unified watertight mesh.

3D-CT scans were resampled to 512x512x512 resolution with 1.0mm isotropic voxels. Using Siddon Ray Tracing \cite{b33, b38}, 512x512 DRR images were generated and enhanced with CLAHE for improved contrast and detail.

\subsection{Bone Localization and Detection}

Anterior-Posterior (AP) DRR images at 512x512 resolution are annotated using Roboflow \cite{b39}. Initial manual annotations establish high-quality labels, followed by the use of Roboflow's autolabeler for efficient dataset annotation. Automatically generated labels are reviewed and corrected to ensure precision. The final dataset is utilized to train the YOLOv9 model, achieving robust and accurate bone detection.

\subsection{Occupancy Value Calculation}

The trimesh Python library ensures watertight meshes for accurate occupancy value calculations \cite{b8}. All meshes are centered and normalized within a unit bounding box. Following Occupancy Networks (ONet) \cite{b8}, \(32^3\) voxel grids are generated. Voxels intersecting the mesh surface are labeled as occupied, and 100,000 random points are analyzed for mesh inclusion using ray-intersection counting. During training of X2B, a subset of 2,048 points is randomly selected to calculate occupancy probabilities, enabling precise 3D reconstructions \cite{b40}.

\section{Experiments}

This section presents the simulation results and comparative analysis of the X2B, and X2BR networks. It provides a comprehensive explanation of the evaluation metrics used and discusses the results in detail. The 3Dr performance of X2BR is compared against three recent single image 2D/3Dr models from the literature: X2V \cite{b26}, D2IM-Net \cite{b12} and ED2IF2-Net \cite{b13}. The section includes an in-depth discussion of the evaluation metrics and a thorough analysis of the findings.

\subsection{Comparative Studies}

This section details the X2V, D2IM-Net and ED2IF2-Net models used for performance comparison.

\subsubsection{X2V}

X2V \cite{b26} reconstructs 3D organ volumes from a single X-ray using an implicit occupancy representation, eliminating the need for templates. Unlike D2IM-Net and ED2IF2-Net, it leverages a Vision Transformer (ViT) encoder for enhanced feature extraction. X2V \cite{b26} achieves superior performance for X-ray to 3D lung volume reconstruction.

\subsubsection{D2IM-Net}

D2IM-Net \cite{b12} is a single-view 3D reconstruction network that combines a coarse implicit field with displacement maps for front and back surfaces to recover topological structures and fine details. Using two decoders to extract global and local features, it achieves high reconstruction quality with metrics like Chamfer Distance (CD), IoU, and Edge Chamfer Distance (ECD). Its Laplacian loss improves surface detail representation \cite{b9}, outperforming DISN.

\subsubsection{ED2IF2-Net}

ED2IF2-Net \cite{b13} leverages Pyramid ViT (PVT) for high-fidelity 3D reconstruction from single RGB images. It disentangles the implicit field into a deformed implicit field for topology and a displacement field for surface details. The architecture features three decoders: a coarse shape decoder, a deformation decoder using implicit field deformation blocks (IFDBs), and a surface detail decoder with hybrid attention modules (HAMs). A comprehensive loss function ensures robust learning of coarse and detailed features.

\subsection{Evaluation Metrics}

The proposed method and baselines are evaluated using Intersection over Union (IoU), Chamfer-L$_1$ (C-L$_1$) distance, F-score, mean absolute \(z\)-axis error (maze), and normal consistency (NC), following the metrics \cite{b26} established in X2V.  

Chamfer-L$_1$ (C-L$_1$) measures the average nearest-point Euclidean distance between predicted and ground truth (GT) meshes. Voxelized IoU quantifies mesh overlap as the intersection-to-union ratio, using a voxelization (with voxel size = 0.5) for the generated meshes. NC evaluates surface normal alignment via the absolute dot product of neighboring points. F-score is the harmonic mean of precision and recall at distance threshold \(t = 0.02\). Mean Absolute \(z\)-axis Error (maze) quantifies alignment deviations in ICP-aligned meshes, with maxe and maye extending this evaluation to x- and y-axes, respectively \cite{b17}. These metrics comprehensively assess reconstruction accuracy and precision \cite{b26}.

\section{Simulations and Results}
\subsection{X2B Performance Analysis}

\begin{figure*}[ht!]
        \centering
        \includegraphics[scale=0.27]{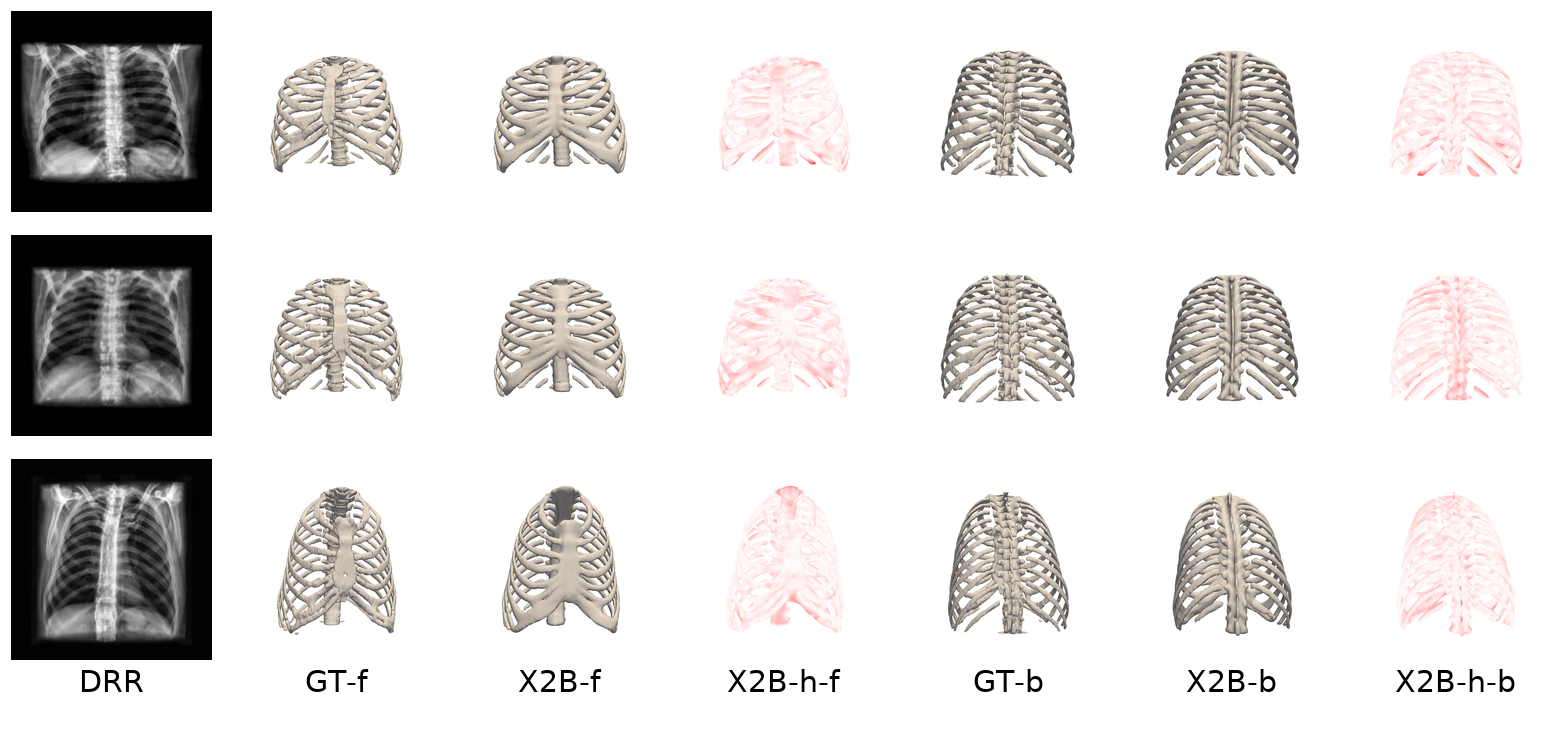}
        \caption{Comparison of X2B and GT. The first column displays the DRR inputs, while the second column presents the GT meshes. The subsequent columns show the X2B reconstructions (X2B-f), X2B heatmaps (X2B-h-f), GT-b, X2B-b, and X2B heatmaps in the back view (X2B-h-b).}
        \label{guven4}
\end{figure*}

Figure \ref{guven4} illustrates X2B reconstructions compared to GT meshes, showing DRR inputs, front and back views of GT and reconstructed meshes, and heatmaps highlighting reconstruction errors. The mean error is 2.192 mm, with a standard deviation of 1.886 mm, reflecting the model’s alignment accuracy and variability. Positive errors in X2B heatmaps result from over-smoothing in occupancy networks, limited training data resolution, and interpolation effects, which bias reconstructions toward overestimated boundaries.

\subsection{X2B \& X2BR Comparison with Existing Methods}

\begin{figure*}[ht!]    
    \centering
    \includegraphics[scale=0.45]{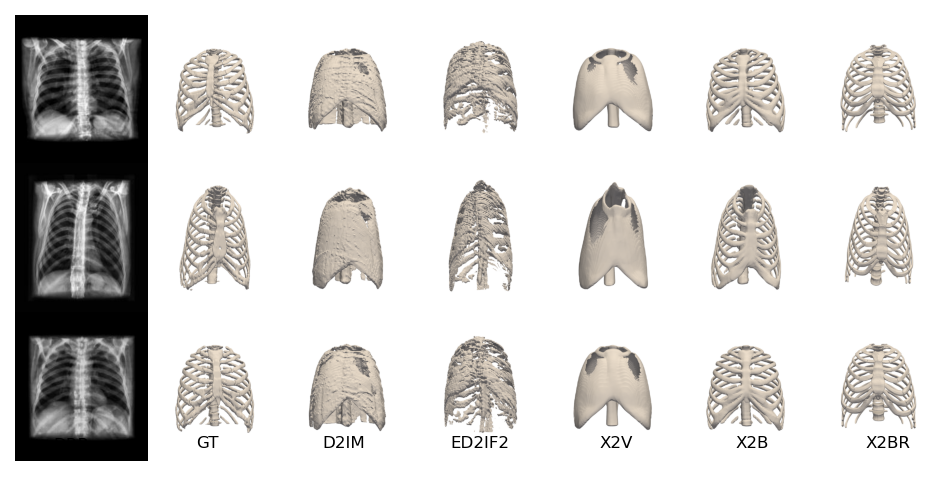}
    \caption{Comparison of reconstruction results across different methods. The figure presents DRR images (leftmost column) and their corresponding GT 3D meshes alongside reconstructed outputs from various methods: D2IM, ED2IF2, X2V, X2B, and X2BR. Each row corresponds to a different DRR input, while the columns illustrate the progression of reconstruction quality across the methods.}
    \label{guven5}
\end{figure*}

Figure \ref{guven5} compares 3Dr results from D2IM-Net, ED2IF2-Net, X2V \cite{b26}, X2B, and X2BR, all trained on the same dataset using DRRs as inputs for a fair evaluation. X2B and X2BR demonstrate superior performance, with X2BR excelling in preserving fine anatomical features, such as rib curvature and spacing, making it particularly suitable for biomedical applications. In contrast, X2V can reconstruct overall organ topology but lacks fine structural details due to the limitations of its implicit representation. D2IM and ED2IF2 struggle with depth variations, overlapping structures, and complex backgrounds, leading to lower reconstruction accuracy.

\begin{table}[h]
\centering
\renewcommand{\arraystretch}{0.9} % Reduce row height
\setlength{\tabcolsep}{3pt} % Reduce column spacing
\caption{Comparison of D2IM, ED2IF2, X2V, X2B, X2BR in terms of IoU, Chamfer-L1, F-score and NC metrics for mesh reconstruction accuracy. }
\begin{tabular}{|c|c|c|c|c|}
\hline
Method & \textit{IoU} & \textit{Chamfer-L1} & \textit{F-score}  & \textit{NC}\\  \hline \hline
\textit{D2IM} & 0.903 $\pm$ 0.028 & 0.009 $\pm$ 0.001 & 0.912 $\pm$ 0.027 & 0.499 $\pm$ 0.006\\ \hline
\textit{ED2IF2} & 0.573 $\pm$ 0.041 & 0.019 $\pm$ 0.002 & 0.567 $\pm$ 0.048 & 0.495  $\pm$ 0.003\\ \hline
\textit{X2V} & 0.859 $\pm$ 0.057 & 0.009 $\pm$ 0.001 & 0.904 $\pm$ 0.050  & 0.496 $\pm$ 0.005 \\ \hline
\textit{X2B} & 0.952 $\pm$ 0.038 & 0.005 $\pm$ 0.001 &  0.974 $\pm$ 0.038  & 0.505 $\pm$ 0.003 \\ \hline
\textit{X2BR} & 0.875 $\pm$ 0.036 & 0.009 $\pm$ 0.001 & 0.913 $\pm$ 0.039 & 0.504 $\pm$ 0.003 \\ \hline

\end{tabular}
\label{guven.t1}
\end{table}

Table \ref{guven.t1} evaluates reconstruction accuracy based on IoU, Chamfer-L1 distance, F-score, and NC. X2B outperforms all methods, achieving the highest IoU, lowest Chamfer-L1 distance, and highest F-score. X2BR performs competitively, with improved NC.

\begin{table}[h]
\centering
\caption{Comparison of D2IM, ED2IF2, X2V, X2B, and X2BR, in terms of maze, maxe and maye metrics in milimeter for mesh reconstruction accuracy.}
\begin{tabular}{|c|c|c|c|}
\hline
Method & \textit{maxe} & \textit{maye} & \textit{maze} \\  \hline \hline
\textit{D2IM} & 3.468 $\pm$ 3.157 & 2.772 $\pm$ 2.474 & 3.304 $\pm$ 3.095 \\ \hline
\textit{ED2IF2} & 11.871 $\pm$ 10.267 & 6.254 $\pm$ 6.655 & 7.114 $\pm$ 6.794 \\ \hline
\textit{X2V} & 4.545 $\pm$ 5.996  & 4.706 $\pm$ 9.352 & 4.442 $\pm$ 6.698 \\ \hline
\textit{X2B} & 2.821 $\pm$ 2.717 & 2.515 $\pm$ 2.320 & 2.590 $\pm$ 2.443 \\ \hline
\textit{X2BR} & 3.562 $\pm$ 3.909 & 3.667 $\pm$ 4.203 & 3.003 $\pm$ 2.793  \\ \hline
\end{tabular}
\label{guven.t2}
\end{table}

Table \ref{guven.t2} compares D2IM, ED2IF2, X2V \cite{b26}, X2B, and X2BR using maxe, maye, and maze metrics. While X2B achieves the highest numerical accuracy, X2BR provides the most visually faithful reconstructions, especially in capturing fine anatomical details such as rib curvature, spacing, and vertebral alignment, as illustrated in Fig. \ref{guven5}. The slightly higher metric errors of X2BR are primarily due to localized deformations introduced during non-rigid registration, which enhance anatomical realism but may slightly misalign with the ground truth mesh in a global coordinate system.

This trade-off reveals a crucial limitation of purely numerical metrics, which can underestimate perceptual or anatomical quality in cases involving fine-grained, patient-specific variations. The template-guided refinement in X2BR enables it to better model true anatomical structure, even if global alignment metrics slightly worsen. Without this qualitative improvement, X2BR’s contribution would be redundant; however, visual comparisons clearly demonstrate its added value in clinical interpretability and anatomical plausibility, which are critical for real-world applications such as surgical planning and biomechanical simulations.

\subsection{Comparison of Registration Techniques}

\begin{figure}[h]    
        \centering
        \includegraphics[width=\columnwidth]{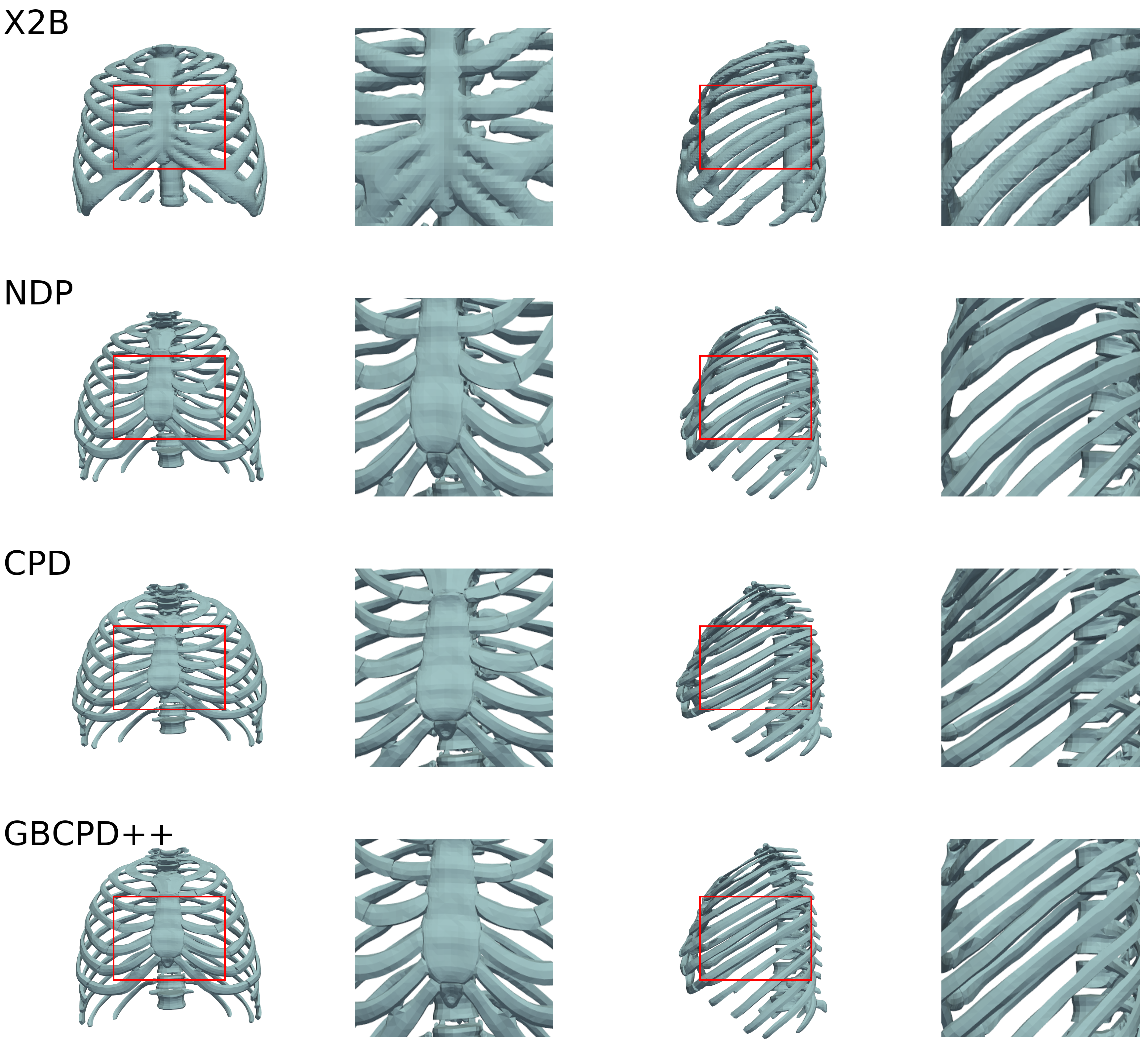}
        \caption{Thorax reconstructions using X2B, NPD, CPD, and GBCPD++. Each row shows full views (front and side) and corresponding zoom-ins highlighting fine anatomical details.}
        \label{guven6}
\end{figure}

\begin{table}[h]
\centering
\renewcommand{\arraystretch}{0.9} % Reduce row height
\setlength{\tabcolsep}{3pt} % Reduce column spacing
\caption{Comparison of GBCPD++, NDP, CPD in terms of IoU, Chamfer-L1, F-score and NC metrics}
\begin{tabular}{|c|c|c|c|c|}
\hline
Method & \textit{IoU} & \textit{Chamfer-L1} & \textit{F-score}  & \textit{NC}\\  \hline \hline
\textit{NDP} & 0.664 $\pm$ 0.111 & 0.008 $\pm$ 0.001 & 0.637 $\pm$ 0.124 & 0.498 $\pm$ 0.003 \\ \hline
\textit{CPD} & 0.744 $\pm$ 0.049 & 0.014 $\pm$ 0.001 & 0.748 $\pm$ 0.051 & 0.498 $\pm$ 0.003 \\ \hline
\textit{GBCPD} & 0.900 $\pm$ 0.019 & 0.010 $\pm$ 0.001 & 0.931 $\pm$ 0.019 & 0.500 $\pm$ 0.004 \\ \hline
\end{tabular}
\label{guven.t3}
\end{table}

Table \ref{guven.t3} and Figure \ref{guven6} evaluate the effectiveness of different non-rigid registration techniques in aligning reconstructed thorax models. The objective is to assess how well each method preserves anatomical structures while handling global and local deformations.  

GBCPD++ achieves the best metrics, including IoU, Chamfer-L1 distance, and F-score, demonstrating superior alignment and structural preservation. CPD shows moderate performance, while NDP exhibits significant misalignments, particularly in rib spacing and structural integrity. These results highlight GBCPD++ as the most effective method for accurate non-rigid registration.

\subsection{Single vs. Double DRRs as Model Input}

This section examines the impact of using single versus bi-planar DRRs on 3Dr accuracy. To address the limitations of a single-view approach, X2B2 extends X2B by integrating two orthogonal DRRs and employing a multi-head cross-attention mechanism for enhanced spatial feature fusion.

\begin{table}[h]
\centering
\caption{Comparison of X2B and X2B2 results in terms of IoU, Chamfer-L1, F-score and NC metrics}
\resizebox{\columnwidth}{!}{%
\begin{tabular}{|c|c|c|c|c|}
\hline
Method & \textit{IoU} & \textit{Chamfer-L1} & \textit{F-score(t=0.02)}  & \textit{NC}\\  \hline \hline
\textit{X2B} & 0.952 $\pm$ 0.038 & 0.005 $\pm$ 0.001 &  0.974 $\pm$ 0.038  & 0.505 $\pm$ 0.003 \\ \hline
\textit{X2B2} & 0.964 $\pm$ 0.020 & 0.004 $\pm$ 0.001 & 0.984  $\pm$ 0.017 & 0.505 $\pm$ 0.003 \\ \hline
\end{tabular}
}
\label{guven.t4}
\end{table}

\begin{table}[h]
\centering
\caption{Comparison of X2B and X2B2 results in terms of maze, maxe, and maye metrics in millimeters}
%\resizebox{\columnwidth}{!}{ % Adjust table to fit within one column
\begin{tabular}{|c|c|c|c|}
\hline
Method & \textit{maxe} & \textit{maye} & \textit{maze} \\ \hline \hline
\textit{X2B} & 2.821 $\pm$ 2.717 & 2.515 $\pm$ 2.320 & 2.590 $\pm$ 2.443 \\ \hline
\textit{X2B2} & 2.473 $\pm$ 2.143 & 2.431 $\pm$ 2.212 & 2.441 $\pm$ 2.247 \\ \hline
\end{tabular}
%}
\label{guven.t5}
%\vspace{-0.3cm}
\end{table}

Tables \ref{guven.t4} and \ref{guven.t5} compare the performance of X2B and X2B2 in 3Dr tasks. Both models perform exceptionally well, but X2B2 demonstrates slight numerical advantages, achieving higher IoU and lower Chamfer-L1 distance , indicating improved geometric accuracy. X2B2 also attains higher F-score, showcasing its robustness in capturing fine structural details, while both models achieve identical NC scores.

Table \ref{guven.t5} highlights X2B2’s improved reconstruction accuracy over X2B in maxe, maye, and maze, showing lower mean errors across all axes. These results reflect X2B2’s ability to reduce positional errors through bi-planar input processing and multi-view data fusion.

\begin{figure}[t]    
        \centering
        \includegraphics[width=\columnwidth]{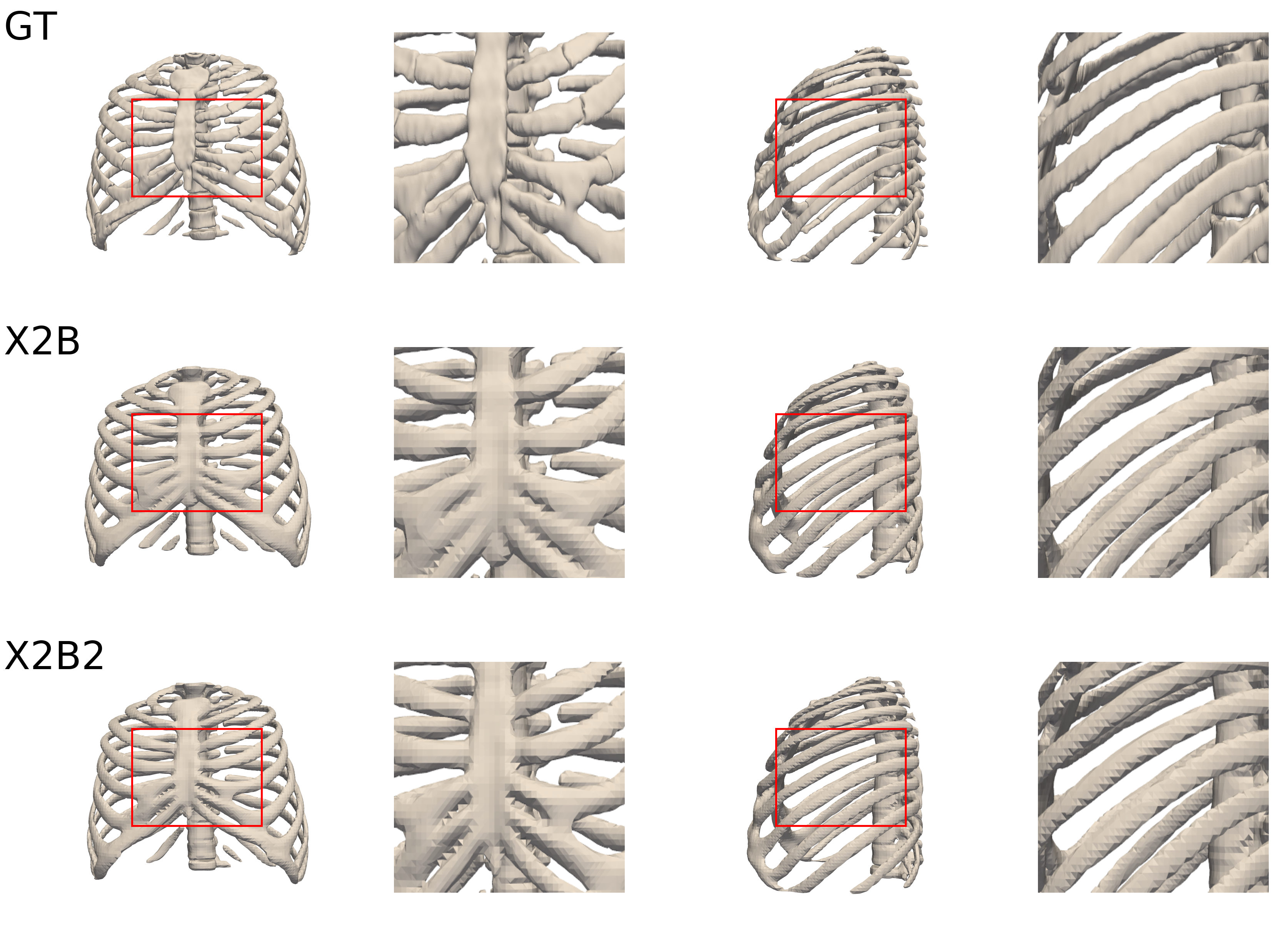}
        \caption{Thorax reconstruction results for GT, X2B, and X2B2. Each row shows full thorax views (front and side) and corresponding zoom-ins to highlight anatomical details.}
        \label{guven7}
\end{figure}

Figure \ref{guven7} compares thorax reconstructions for the GT, X2B, and X2B2 models. The zoomed-in regions emphasize X2B2’s advantage in capturing finer anatomical details, illustrating its incremental improvement over X2B in reconstructing complex thorax structures.

X2B2 shows a slight numerical advantage over X2B, particularly in preserving fine anatomical details in 3Drs. By leveraging bi-planar DRR inputs and cross-attention, it enhances spatial feature fusion, improving reconstruction accuracy. These refinements make X2B2 well-suited for medical imaging and multi-perspective 3D reconstruction while maintaining X2B’s efficiency.

\section{Conclusion}

This study presents X2B and X2BR, two complementary neural implicit frameworks for high-fidelity 3D skeletal reconstruction from a single planar X-ray. X2B leverages a ConvNeXt-based encoder and Conditional Batch Normalization (CBN) layers to predict continuous occupancy fields, enabling template-free reconstruction of complex anatomical structures such as ribs and vertebrae. X2BR builds upon this by integrating a biomechanical template and applying non-rigid registration through GBCPD++, enhancing anatomical plausibility and improving alignment to patient-specific skeletal morphology.

Both frameworks capitalize on the strengths of neural implicit representations—modeling continuous volumetric structures without reliance on voxel grids—while addressing key challenges in X-ray-based reconstruction, including occlusion, overlapping intensities, and incomplete input data. Evaluations on clinical datasets demonstrate that X2B achieves state-of-the-art accuracy in metrics such as volumetric IoU, Chamfer-L1 distance, and F-score, while X2BR offers improved anatomical consistency through template-guided refinement.

In addition to methodological advances, this work introduces the largest known dataset of paired 3D bone meshes and corresponding digitally reconstructed radiographs (DRRs), contributing a valuable benchmark for future research. By addressing long-standing limitations in 3D reconstruction from sparse imaging data, X2B and X2BR offer practical tools for surgical planning, orthopedic assessment, and personalized biomechanical simulations.

%\appendices

%\section*{References and Footnotes}

%\subsection{References}

\end{document}